\documentclass[sigconf,natbib=true]{acmart}
\usepackage{multirow}
\usepackage{tabularx}
\usepackage{bbding}
\usepackage{utfsym}
\usepackage{extarrows}
\usepackage{color} 
\usepackage{xcolor}
\usepackage{makecell}
\usepackage{amsmath,amsfonts}
\usepackage{algorithmic}
\usepackage{array}
\usepackage{graphicx}
\usepackage{stfloats}
\usepackage{multicol}
\usepackage{multirow}
\usepackage{hhline}
\usepackage{colortbl}
\usepackage{bm}
\usepackage{url}
\usepackage{enumitem}
\usepackage{arydshln} 
\usepackage{fontawesome5} 

\AtBeginDocument{%
  }

\copyrightyear{2025}
\acmYear{2025}
\setcopyright{acmlicensed}\acmConference[SIGIR '25]{Proceedings of the 48th
International ACM SIGIR Conference on Research and Development in
Information Retrieval}{July 13--18, 2025}{Padua, Italy}
\acmBooktitle{Proceedings of the 48th International ACM SIGIR Conference on
Research and Development in Information Retrieval (SIGIR '25), July 13--18,
2025, Padua, Italy}
\acmDOI{10.1145/3726302.3729979}
\acmISBN{979-8-4007-1592-1/2025/07}
\begin{document}

\title{FiRE~\textcolor{red}{\faFire} : Enhancing MLLMs with Fine-Grained Context Learning for Complex Image Retrieval}

\author{Bohan Hou}
\affiliation{%
  \institution{Shandong University}
 \city{Qingdao}
 \state{Shandong}
 \country{China}
}
\email{bohanhou@foxmail.com}

\author{Haoqiang Lin}
\affiliation{%
  \institution{Shandong University}
\city{Qingdao}
 \state{Shandong}
 \country{China}
}
\email{zichaohq@gmail.com}
\author{Xuemeng Song*}
\affiliation{%
  \institution{City University of Hong Kong}
\city{Hong kong}
 \country{China}
}
\email{sxmustc@gmail.com}
\author{Haokun Wen}
\affiliation{%
  \institution{Harbin Institute of Technology (Shenzhen)}
 \city{Shenzhen}
 \state{Guangdong}
 \country{China}
 }
\email{whenhaokun@gmail.com}

\author{Meng Liu}
\affiliation{%
  \institution{Shandong Jianzhu University }
 \city{Jinan}
 \state{Shandong}
 \country{China}
  }
\email{mengliu.sdu@gmail.com}

\author{Yupeng Hu}
\affiliation{%
  \institution{Shandong University }
  \city{Jinan}
 \state{Shandong}
  \country{China}
 }
\email{huyupeng@sdu.edu.cn}

\author{Xiangyu Zhao}
\affiliation{%
  \institution{City University of Hong Kong}
 \city{Hong kong}
 \country{China}
  }
\email{xy.zhao@cityu.edu.hk}
\renewcommand{\shortauthors}{Bohan Hou et al.}

\thanks{*Xuemeng Song (sxmustc@gmail.com) is the corresponding author.}
\begin{abstract}
Due to their strong generalizable multimodal processing and reasoning capabilities, Multimodal Large Language Models (MLLMs) have demonstrated significant potential as universal image retrievers, effectively addressing diverse real-world image retrieval tasks. Nevertheless, pioneering studies, while promising, overlook the potential of fine-grained context modeling and disentangled fine-tuning objectives in enhancing MLLMs' retrieval performance, particularly for complex tasks such as long-text-to-image retrieval, visual dialog retrieval, and composed image retrieval (CIR).
Therefore, in this work,  we propose an automated fine-grained multimodal quintuple dataset construction pipeline and a novel two-stage fine-grained multimodal fine-tuning strategy.  The dataset generation pipeline produces a comprehensive CIR dataset with fine-grained image captions and modification text, facilitating fine-grained context modeling. Beyond the previously entangled fine-tuning paradigm, our approach separates the fine-tuning process into two distinct stages: (1) fine-grained context reasoning-oriented fine-tuning and (2) fine-grained retrieval-oriented fine-tuning. These stages aim to sequentially enhance the model's context understanding and query-target alignment capabilities, thereby improving retrieval performance.
Extensive experiments across five datasets encompassing diverse and complex image retrieval tasks demonstrate the remarkable superiority of our method over existing approaches in zero-shot retrieval settings, even with a more lightweight MLLM backbone compared to those methods. 

\end{abstract}



\maketitle

\section{Introduction}

To meet the diverse demands of users in real-world applications~\cite{scss,fashioniq,imageret-app,song2017neurostylist}, various image retrieval paradigms have been proposed. These include the standard short-text-to-image retrieval~\cite{mscoco,flickr,crossmodal}, where a short caption suffices to express the user's search intent, as well as more complex paradigms like long-text-to-image retrieval~\cite{long-clip}, dialog-based image retrieval~\cite{visualdialog}, and composed image retrieval (CIR)~\cite{tirg,sac}. In these more challenging scenarios, the user's complex search intent often needs to be conveyed through long, detailed queries or composed multimodal inputs (i.e., a reference image paired with a modification text), significantly increasing retrieval complexity.
Existing approaches typically address these tasks independently, resulting in high training costs and fragmented solutions. To improve efficiency and scalability, recent research has shifted toward developing unified retrieval models capable of handling diverse tasks and query types within a single framework.

\begin{figure}[t]
		\centering
		\includegraphics[width =\linewidth]{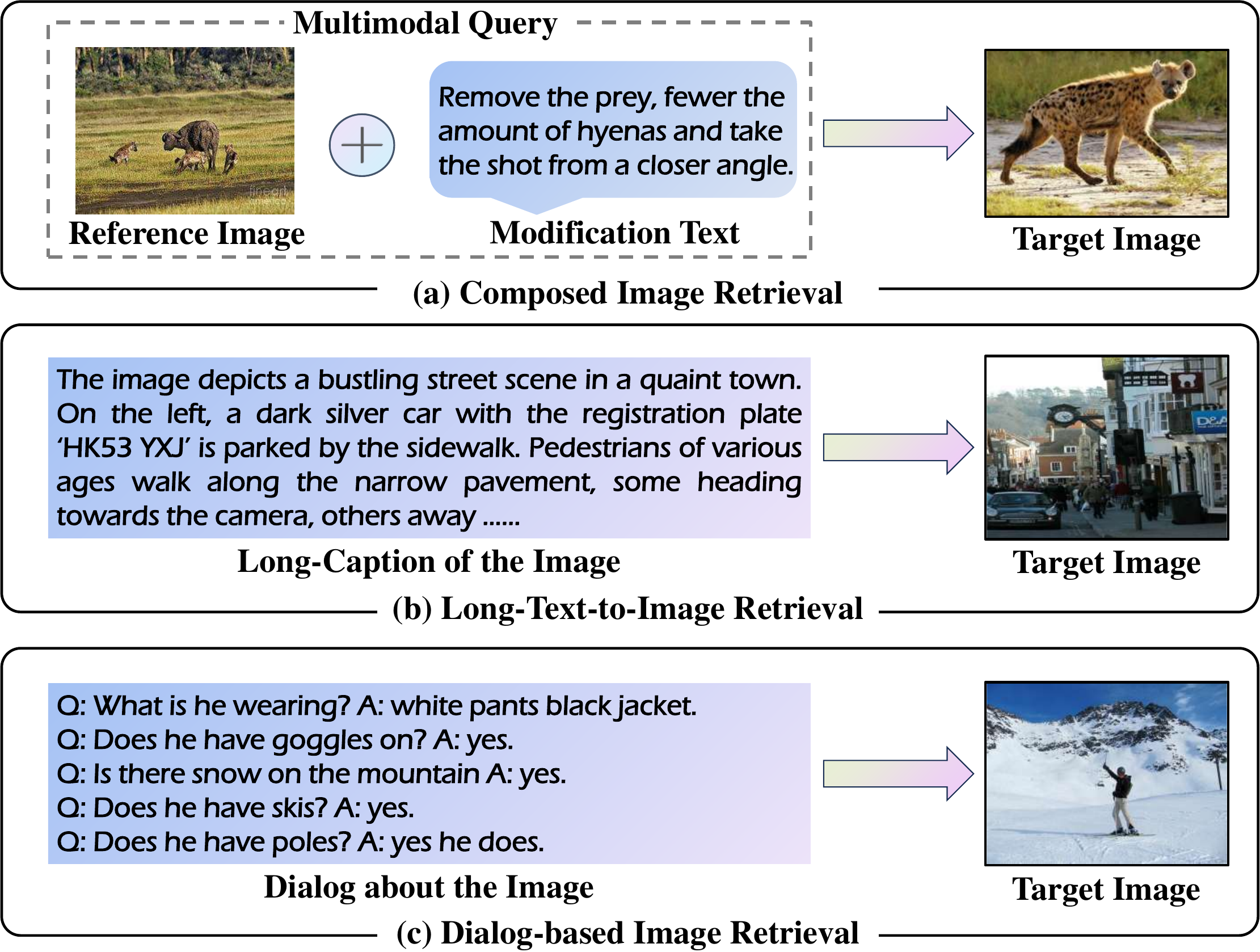}
	    \caption{
        Illustration of complex image retrieval tasks: (a) Composed Image Retrieval, (b) Long-Text-to-Image Retrieval, and (c) Dialog-based Image Retrieval. 
        }
        \label{Intro}
        \vspace{-1.5em}
\end{figure}

Pioneer works~\cite{magiclens,clip4cir} typically rely on Vision-Language Pretraining Models (VLPs)~\cite{clip,blip2}, utilizing their strong multimodal embedding capabilities to address various retrieval tasks. However,  VLPs struggle to understand complex queries~\cite{lincir}, particularly those involving reasoning, due to their relatively small model scale. 
Recent studies have turned to Large Language Models (LLMs), which offer superior language understanding and reasoning capabilities~\cite{e5v}, to address diverse image retrieval tasks. Since LLMs, as generative models, inherently lack the discriminative query-target alignment capabilities needed for retrieval tasks~\cite{finetunellama,grid,frogma}, these studies focus on designing effective fine-tuning strategies to bridge this gap. For instance, MCL~\cite{mcl} fine-tunes an LLM, integrated with a CLIP visual encoder and an adaptor, on two tasks—multimodal-context captioning and multimodal-context retrieval—using a custom large-scale multimodal composition dataset. Conversely, E5-V~\cite{e5v} fine-tunes the core LLM component of a pre-trained MLLM on pure sentence pairs, aiming to enhance its query-target alignment capabilities.

Despite promising results, existing methods face two key limitations that hinder their performance on complex image retrieval tasks. \textbf{1) Lacking fine-grained context modeling}.  Existing methods rely on sentence pairs or \textless \emph{reference image, short modification text, short target caption}\textgreater ~triplets to fine-tune MLLMs. However, these data types provide only coarse-grained information and lack the detailed descriptions necessary for developing the model's ability to fine-grained context understanding. In fact, queries in many real-world retrieval tasks often involve long or composite contexts (see Figure~\ref{Intro}), which inherently demand fine-grained comprehension for accurate interpretation. Moreover, modern MLLMs typically represent each image as extensive token embeddings, akin to processing long contexts, further highlighting the importance of fine-grained context modeling.

\textbf{2) Suboptimal fine-tuning objective.} 
E5-V's fine-tuning strategy focuses solely on improving the MLLM's homogeneous query-target alignment capability while neglecting the optimization of its multimodal context understanding ability.
Although MCL incorporates multimodal context learning, it fine-tunes MLLMs simultaneously on both multimodal retrieval and generation tasks. This entangled fine-tuning objective may compromise the balance between enhancing context understanding and improving query-target alignment, ultimately resulting in suboptimal retrieval performance.

Accordingly, our goal is to address these limitations by enhancing the MLLM with fine-grained context learning capabilities, transforming it into a powerful universal image retriever capable of handling various image retrieval tasks, particularly complex ones. Specifically, to address the first limitation, we propose a fully automated pipeline for generating a large-scale fine-grained multimodal composition dataset. The pipeline consists of three stages: (1) CoT-based fine-grained caption generation, where an LLM is guided to produce detailed captions through fine-grained reasoning steps, including subject-oriented, attribute-oriented, and context-oriented reasoning; (2) MLLM-based image pair identification, which uses fine-grained semantic similarity—rather than conventional visual or coarse-grained similarity—to identify potential reference-target image pairs. This stage involves fine-tuning an MLLM with \textless \emph{image, fine-grained caption}\textgreater ~pairs to enhance its long-text encoding capability; and (3) human-like fine-grained modification generation, where a vagueness-guided instruction is designed to bridge discrepancies between LLM outputs and human annotations.

Using this pipeline, we create a large-scale Fine-Grained Multimodal Quintuple dataset, named \textbf{FiGMaQ}, with 87K samples, where each quintuple sample follows the format \textless \emph{reference image, reference caption, modification text, target image, target caption}\textgreater. Compared to existing LLM-generated multimodal triplet datasets, our dataset offers three significant features. 1) Image captions are fine-grEained, containing detailed information with an average of over 100 tokens. 2) Modification text captures more specific details and is more human-like, accommodating vague terms.
3) Each sample includes five components, facilitating a variety of fine-tuning tasks, such as multimodal-context captioning and retrieval.

To address the second limitation, we propose a two-stage Fine-gRained multimodal finE-tuning strategy, named FiRE, which aims to sequentially strengthen the MLLMs' context understanding and query-target alignment capabilities with disentangled fine-tuning objectives. Specifically, inspired by the success of CIR-based fine-tuning in boosting MLLM performance on various zero-shot retrieval tasks~\cite{mcl}, we adopt CIR—requiring complex contextual understanding and fine-grained reasoning—as the representative task for MLLM fine-tuning. 
In the first stage, we perform fine-grained context reasoning-oriented fine-tuning by instructing MLLMs to generate fine-grained captions for target images based on reference images and modification text.
In the second stage, building on the improved multimodal reasoning capability developed in the first stage, we conduct fine-grained retrieval-oriented fine-tuning, where both InfoNCE loss and Recall@$k$ surrogate loss are used to boost the MLLM's query-target alignment capabilities.

In summary, our contributions can be summarized as follows:
\begin{itemize}
\item We propose an automated pipeline for generating fine-grained multimodal composition datasets and contribute a large-scale dataset, FiGMaQ, to support future research on fine-grained context learning with MLLMs.
\item We propose a two-stage fine-grained fine-tuning approach that separately strengthens the MLLM's abilities in complex context learning and query-target alignment, promoting its adaptation to diverse image retrieval tasks. With disentangled fine-tuning objectives, our fine-tuning approach requires fewer computational resources.

\item We conducted extensive experiments on seven datasets spanning diverse image retrieval tasks, demonstrating the remarkable superiority of our method—achieving state-of-the-art performance across various complex image retrieval tasks in the zero-shot one-checkpoint setting—even with a more lightweight MLLM backbone.
\end{itemize}

\begin{figure*}[!t]
    \centering
    \includegraphics[width=\linewidth]{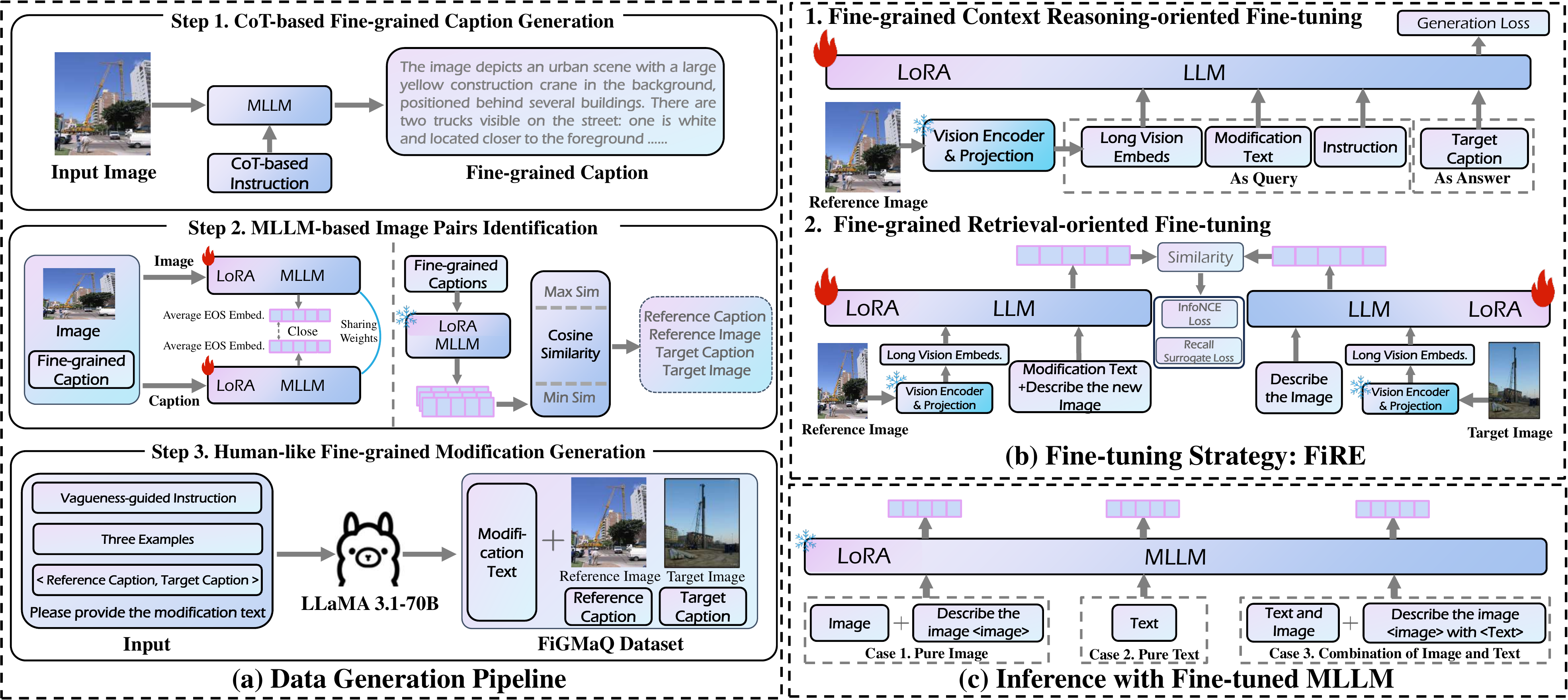}
    
    \caption{Illustration of Our Method. (a) illustrates the data generation pipeline, (b) presents our fine-tuning strategy: FiRE, and (c) provides inferencing with fine-tuned MLLM.
}\label{fig:model_struct}
\vspace{-1em}
\end{figure*}
\vspace{-1em}
\section{Related Work}
\textbf{Multimodal Composition Dataset}. 
Existing works~\cite{mcl} have demonstrated  multimodal composition data can enhance MLLMs' understanding of multimodal inputs. Such data typically resembles CIR samples, taking the triplet form \textless \emph{reference image, modification text, target image}\textgreater. Early CIR datasets~\cite{cirr,fashioniq,searle} are mainly manually annotated, making them limited in scale due to prohibitively expensive annotation costs. To address this issue, researchers have proposed several automated triplet generation methods, which fall into two categories. 
1) \textbf{Semi-Automated Annotation.} This strategy~\cite{case} uses LLMs to annotate modification text for image pairs but relies on human intervention for image selection and triplet evaluation. For example, LaSCo~\cite{case} builds its CIR dataset from the VQA2.0~\cite{vqa} dataset. Initially, it provides human volunteers with 24 visually similar images for each image, allowing them to select relevant images to form image pairs. Subsequently, it feeds the QA information of related image pairs into LLMs to generate modification text and employs human reviewers to assess the quality of the generated triplets. While this approach reduces manual annotation costs, it remains expensive and hard to scale due to human involvement. 
2) \textbf{Fully Automated Annotation.} 
This strategy~\cite{mcl} aims to fully automate triplet data generation without human intervention. For example, MCL~\cite{mcl} generates its MMC dataset by using LLMs to produce modification and target captions based on a reference image and its derived caption. However, the absence of real target images results in low-quality triplets.
In contrast, MagicLens~\cite{magiclens} automatically identifies potential reference-target image pairs by first grouping images from the same webpage,  then generating metadata for each image with various annotation tools, and finally using where image both CLIP-based visual similarity and textual similarity are used for potential pairs filtering. The metadata of these image pairs is then processed by an LLM to generate modification text. A major limitation of this approach is that the generated metadata remains coarse-grained, capturing only general and common attributes of the main subjects in the images. To address this issue, we propose a fully automated pipeline for producing fine-grained multimodal composition datasets.

\vspace{-1em}
\section{\textbf{F}ine-\textbf{G}rained \textbf{M}ultimod\textbf{a}l  Dataset Generation}

In this section, we present our fine-grained multimodal quintuple dataset generation pipeline, as shown in Figure~\ref{fig:model_struct}(a), consisting of three stages: CoT-based fine-grained image caption generation, MLLM-based image pairs identification, human-like fine-grained modification generation. 

\subsection{CoT-based Fine-grained Caption Generation.}
In this stage, we aim to generate the images' fine-grained captions, providing detailed context as the input for the subsequent MLLM-based potential query-target image pairs identification. For this purpose, we utilize the unlabeled test split of ImageNet1K~\cite{imagenet}, comprising 100K unlabeled, open-domain, real-world images featuring a wide variety of subjects, as the initial dataset.

\begin{figure}[t]
		\centering
		\includegraphics[width = \linewidth]{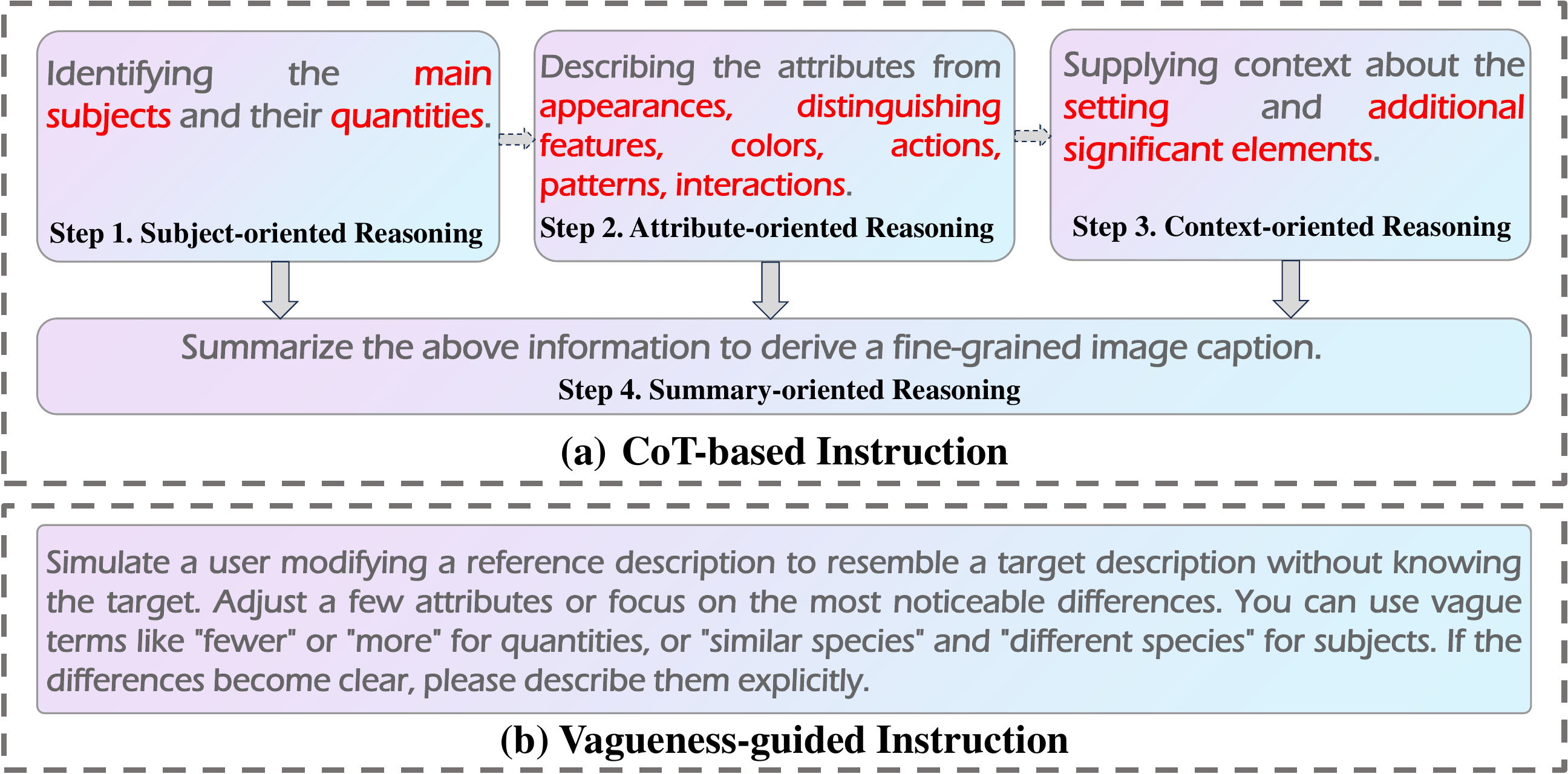}
	    \caption{Illustration of instructions involved in: (a) CoT-based Instruction and  (b) Vagueness-guided Instruction.
}\label{prompt}
\vspace{-1.5em}
\end{figure}

Instead of requiring the MLLM to generate a fine-grained image caption in a single reasoning step, we design a CoT instruction that encourages the generation of detailed captions through a sequence of fine-grained reasoning steps, ensuring richness and specificity. Intuitively, humans process visual information in stages—starting with the identification of principal subjects, followed by noticing their fine-grained attributes, and finally considering the background context.
Accordingly, as shown in Figure~\ref{prompt}(a), we structure the CoT-guided instruction into four key reasoning steps: subject-oriented reasoning, attribute-oriented reasoning, context-oriented reasoning, and summary-oriented reasoning. 
The first step directs the MLLM to identify the subjects and their quantities, while the second step focuses on outputting the detailed attributes of these subjects. In this work, we define six types of attributes—appearance, color, pattern, distinguishing features, action, and interaction—to guide the MLLM in capturing the subjects' detailed properties. Next, in the context-oriented reasoning step, we guide the model to provide contextual information, including the setting and other significant elements. Finally, the summary-oriented reasoning step synthesizes the outputs from the previous three steps to form a coherent and comprehensive fine-grained image caption.
Let \( D_I \) denote the generated fine-grained caption for each original image \( I \), which is a relatively long text, averaging over 100 tokens. 

\subsection{MLLM-based Image Pairs Identification}
Having obtained fine-grained image-caption pairs, we proceed to create the quintuple sample. First, we need to identify relevant image pairs to form reference-target pairs, and then generate modification text for each pair. Unlike previous CIR dataset generation methods~\cite{magiclens, cirr, case} that primarily rely on visual or coarse-grained similarity, we use fine-grained semantic similarity for selecting relevant image pairs. This approach is motivated by our observation that visually similar image pairs can often be semantically irrelevant (e.g., images with a similar visual style but unrelated content)
, which does not align with practical user retrieval demands.

Considering that the existing VLP encoder~\cite{lincir} struggles with excessively fine-grained texts (e.g., our fine-grained captions), we turn to an LLM to leverage its strong context comprehension capabilities. However, since the LLM is a generative model, not originally designed for retrieval tasks, we propose fine-tuning it to improve its fine-grained long-text encoding capability. To achieve this, we utilize the fine-grained image-caption pairs obtained in Subsection 3.1 to optimize an MLLM for the task of image-text alignment. We then employ the LLM component of the fine-tuned MLLM for long-text encoding, enabling semantic similarity assessment for relevant image pair identification.

Specifically, inspired by the decoder-only LLM-based query-document retrieval model~\cite{finetunellama}, which appends an End-of-Sequence (EOS) token to the end of a given token sequence to summarize its semantic content, we append $M$ EOS tokens to each image token sequence and its corresponding caption token sequence. We then average the $M$ EOS token embeddings from the MLLM's last hidden states to obtain the final representation of the input image/caption. This process can be formulated as follows:
\begin{equation}
    \mathbf{f}
= \frac{1}{M}
\sum_{m=1}^{M}
\mathrm{LLM}\bigl(\mathrm{Input} ; M \cdot \mathrm{[EOS]}\bigr)\bigl[-m\bigr].
\label{eq:feature}
\end{equation} 
Thereafter, we employ the commonly used image-text contrastive loss for fine-tuning, which can be formalized as follows:
\begin{equation}
\begin{aligned}
\mathcal{L}_{align}
= - \frac{1}{2B} \sum_{i=1}^{B} 
\Biggl[
    \log \frac{\exp\bigl(s_{ii} / \tau\bigr)}
              {\sum_{j=1}^{B} \exp\bigl(s_{ij} / \tau\bigr)}
    \;+\;
    \log \frac{\exp\bigl(s_{ii} / \tau\bigr)}
              {\sum_{j=1}^{B} \exp\bigl(s_{ji} / \tau\bigr)}
\Biggr].
\end{aligned}
\label{eq:match_Loss}
\end{equation}
Here, $s_{ij}= \cos\langle \mathbf{f}_i^v, \mathbf{f}_j^t\rangle$ represents the similarity between image $i$ and text $j$. $B$ is the batch size, and $\tau$ is a temperature parameter. 

Once the MLLM is adequately trained using the above loss function, it can be employed to identify potential image pairs. To exclude overly similar or irrelevant image pairs, which would not be useful for practical modification tasks, we adopt thresholds for cosine similarity, as suggested in~\cite{cirr, magiclens}. Specifically, we define an upper threshold  $\theta_h$ and a lower threshold  $\theta_l$, and only retain image pairs whose cosine similarities lie within this range, as follows: 
\begin{equation}
    \{ \langle I_r, I_t \rangle \;|\; \cos\langle \mathbf{f}_{{I_r}}, \mathbf{f}_{{I_t}} \rangle \in [\theta_l, \theta_h] \}.
\label{related_pairs}
\end{equation}

\subsection{Human-like Fine-grained Modification Generation}
Having obtained relevant image pairs, we can proceed to automated modification text generation.  Existing works~\cite{magiclens,transagg} typically directly prompt LLMs to generate modification texts based on coarse-grained information (e.g., global captions) of image pairs. However, this approach has two key limitations: 1) hindering modification of fine-grained attributes due to the lack of detailed inputs, and  2) the generated modification texts tend to be overly precise, e.g., ``change fifteen people to seven'' and ``replace dark red to light red'',  due to LLM's powerful reasoning capability. This precision deviates from real-world users' vaguer expressions (e.g., ``fewer people'', and ``lighter color'') when making modification requests.

To facilitate fine-grained modifications and ensure natural expression, we propose a human-like fine-grained modification generation scheme. Unlike previous studies, we feed the fine-grained captions of image pairs into an LLM, enabling it to generate more accurate modification texts. Additionally, we design a vagueness-guided instruction to encourage LLMs to produce human-like modifications. As shown in Figure~\ref{prompt}(b), our instruction simulates real-world scenarios and guides the model to generate vague modifications. Notably, we do not enforce the LLM to always produce vague modifications, but allow it to provide precise changes when the modification demands are highly specific (e.g., ``changing a cat into a dog'').
To further enhance the LLM's understanding of this instruction, we provide three human-annotated examples that encompass both vague modifications and direct comparisons, leveraging the LLM's robust few-shot learning capabilities~\cite{llm-is-few-shot-learner}.
 To ensure the quality of the generated modification text, we use a strong LLM, specifically LLaMA 3.1 $70$B\footnote{https://ai.meta.com/blog/meta-LLaMA-3-1/}.
Ultimately, through this data generation pipeline, we construct a dataset of approximately 87K scalable quintuplets, each containing \textless \emph{reference image, fine-grained reference caption, modification text, target image, fine-grained target caption}\textgreater.
\section{Fine-grained Multimodal Fine-tuning}

In this section, we present our two-stage fine-tuning strategy, as shown in Figure~\ref{fig:model_struct}(b), comprising fine-grained context reasoning-oriented fine-tuning and fine-grained retrieval-oriented fine-tuning.

\subsection{Fine-grained Context Reasoning-oriented Fine-tuning}

MLLMs inherently lack sufficient fine-grained context modeling capability, resulting in deficiencies in fine-grained context reasoning and limiting their effectiveness in complex tasks. To overcome this limitation, we perform fine-grained context reasoning-oriented fine-tuning using instruction tuning~\cite{instructiongpt,instruction}. Specifically, similar to~\cite{mcl}, we construct instruction-answer pairs based on our generated multimodal quintuples.
The instruction is formatted as: ``\textit{<reference image> but modified with modification text. Please describe the new image}'', while the answer corresponds to the \textit{fine-grained target caption}.
Notably, different from~\cite{mcl} which simply uses a single CLIP-based visual embedding, we use the long sequence of token embeddings yielded by MLLMs for representing the \textit{<reference image>}.  This approach is based on the idea of treating the reference image as analogous to a long text, offering two key benefits:  1) encouraging the MLLM to develop fine-grained context reasoning, and 2) enhancing the MLLM's generalization to other complex image retrieval tasks, such as long-text-to-image and dialog-based image retrieval. 
We fine-tune the MLLM with a generative loss, which can be formalized as follows:
\begin{equation} 
    \mathcal{L}_{gen} = -\sum_{i=1}^{N} \log P_{LLM}(x_i|\Phi, x_{1:{i-1}}),
    \label{eq: loss_gen}
\end{equation}
where \(\Phi\) represents the embeddings of the instruction, \(N\) denotes the length of the token embeddings of the answer, and \(x_i\) represents the \(i\)-th token embedding of the answer.

\subsection{Fine-grained Retrieval-oriented Fine-tuning}
This stage aims to optimize the MLLM's query-target alignment capability, building on the MLLM whose multimodal reasoning capability has been enhanced from the first stage. Specifically, similar to MCL~\cite{mcl}, we adopt the composed image retrieval task for fine-tuning. However, different from MCL which targets aligning a multimodal query feature to an unimodal target caption feature, we propose aligning the multimodal query feature to a multimodal target feature. 
In particular, we format the query as \textit{``\(\langle\text{reference image}\rangle\) but \textbf{modification}, please describe the new image''} and the target as \textit{`` \(\langle\text{target image}\rangle\) describe the image''}. Adding the prompt ``describe the image'' leads to a unified multimodal input format, facilitating the cross-modal query-target alignment. 

Specifically, we derive the multimodal query and target features following Eq.(\ref{eq:feature}) and use the batch-based InfoNCE loss for query-target alignment optimization, which can be formalized as:
\begin{equation}
\mathcal{L}_{{InfoNCE}} = -\frac{1}{B}\sum_{i=1}^{B} \log \frac{\exp\left(\cos\langle \mathbf{f}^\text{query}_i, \mathbf{f}^\text{target}_i \rangle / \tau'\right)}{\sum_{j=1}^{B} \exp\left(\cos\langle \mathbf{f}^\text{query}_i, \mathbf{f}^\text{target}_j \rangle / \tau'\right)},
\label{eq:infonce_loss}
\end{equation}
where $B$ is the batch size, and $\tau'$ is a temperature parameter.

To further enhance the MLLM's ability to achieve discriminative query-target alignment, we introduce the Recall@$k$ Surrogate Loss~\cite{recallloss}. This loss function directly constrains the rank of the ground truth target within the same batch, effectively improving target retrieval performance. The formula is as follows: 
\begin{equation}
\left\{
\begin{aligned}
&\tilde{R}^{k}_{\Omega}(q) = \frac{\sum_{x \in \mathcal{P}_q} \sigma_{\tau_1}\bigl(k - 1 - \sum\limits_{z \in \Omega, z \neq x}\sigma_{\tau_2}(s_{qz} - s_{qx})\bigr)}{|\mathcal{P}_q|}, \\
&\mathcal{L}_{recall}^{k} = \frac{1}{B}\sum_{i=1}^{B}(1 - \tilde{R}^{k}_{\Omega}(q_i)),
\end{aligned}
\right.
\label{eq: recall loss}
\end{equation}
where $s_{qz}$ represents the similarity score between the query $q$ and a candidate $z$, $\mathcal{P}_q$ represents the set of ground-truth (positive) matches for the query $q$, and $\Omega$ represents the set of all candidate items. $\sigma_{\tau_1}$ and $\sigma_{\tau_2}$ represent sigmoid functions with temperature parameters $\tau_1$ and $\tau_2$, respectively. $\tilde{R}^{k}_{\Omega}(q)$ is the differentiable Recall@$k$ surrogate for a given query $q$. 

In summary, our total loss can be written as:
\begin{equation}
\mathcal{L}_{Total} = \mathcal{L}_{{InfoNCE}} + \sum_{k \in \mathcal{R}_s} \beta_{k} \mathcal{L}_{{Recall}}^{k},
\label{eq:total_loss}
\end{equation}
where $\mathcal{R}s$ is the set of $k$'s adopted for recall optimization. $\{\beta_{k}\}$ are hyperparameters that control the contributions of adopted recall surrogate losses. 

\subsection{Inference with the Fine-tuned MLLM}
During inference, the fine-tuned MLLM encodes the query and candidate target images separately. The target image is then retrieved based on its similarity to the query. As illustrated in Figure~\ref{fig:model_struct}(c), the fine-tuned MLLM supports various query types, including pure text, pure image (using the instruction \textit{describe the image <image>}''), and a combination of image and text (using the instruction \textit{describe the image <image> with <text>}''), where <image> and <text> are replaced by the corresponding image and text token embeddings.

\section{Experiment}
In this section, we first introduce the experimental settings and then provide the experiment results.

\subsection{Experiment Settings}
\subsubsection{Evaluation Dataset.} To comprehensively evaluate the effectiveness of our method across various image retrieval tasks, we selected three complex retrieval tasks: CIR, long-text-to-image retrieval, and dialog-based image retrieval, as well as a simpler task, i.e., short-text-to-image retrieval.
For the CIR task, we adopted three commonly used datasets, including two open-domain datasets: \textbf{CIRR} dataset~\cite{cirr} and \textbf{CIRCO} dataset~\cite{searle}, as well as a fashion domain dataset: \textbf{FashionIQ}, which can be further divided into three subsets: Dresses, Shirts, and Tops\&Tees.
For long-text-to-image retrieval, we adopted the publicly available \textbf{Urban1K} dataset~\cite{long-clip}, which consists of 1K images of cityscapes with subtle differences, and each image is accompanied by a fine-grained description. For dialog-based image retrieval, we used the \textbf{Visual Dialog} dataset~\cite{visualdialog}. For short-text-to-image retrieval, we selected two classic datasets: \textbf{COCO}~\cite{mscoco} and \textbf{Flickr30K}~\cite{flickr}.

\begin{table*}[!t]
    \centering
    \caption{
        \textbf{Performance comparison on CIRR in terms of
        $R@k$ (\%) and $R_{\mathrm{subset}}@k$ (\%), and on CIRCO
        in terms of $\mathrm{mAP}@k$ (\%).}
        The best results are shown in bold, and -- denotes unavailable
        results. We also report the absolute performance improvements
        of our model over dedicated ZS-CIR models and universal retrieval
        models.
    }
    \label{tab:baseline_cirr}

    \resizebox{\linewidth}{!}{%
        \begin{tabular}{l|cccc|ccc|cccc}
            \hline
            \multirow{3}{*}{Method}
            & \multicolumn{7}{c|}{CIRR}
            & \multicolumn{4}{c}{CIRCO} \\
            \cline{2-12}

            & \multicolumn{4}{c|}{$R@k$}
            & \multicolumn{3}{c|}{$R_{\mathrm{subset}}@k$}
            & \multicolumn{4}{c}{$\mathrm{mAP}@k$} \\
            \cline{2-5}
            \cline{6-8}
            \cline{9-12}

            & $k=1$ & $k=5$ & $k=10$ & $k=50$
            & $k=1$ & $k=2$ & $k=3$
            & $k=5$ & $k=10$ & $k=25$ & $k=50$ \\
            \hline
            \hline

            Pic2Word~\cite{pic2word}
            {\footnotesize\textcolor{gray}{(CVPR'23)}}
            & 23.90 & 51.70 & 65.30 & 87.80
            & -- & -- & --
            & 8.72 & 9.51 & 10.46 & 11.29 \\

            LinCIR~\cite{lincir}
            {\footnotesize\textcolor{gray}{(CVPR'24)}}
            & 25.04 & 53.25 & 66.68 & --
            & 57.11 & 77.37 & 88.89
            & 12.59 & 13.58 & 15.00 & 15.85 \\

            SEARLE-XL-OTI~\cite{searle}
            {\footnotesize\textcolor{gray}{(ICCV'23)}}
            & 24.87 & 52.31 & 66.29 & 88.58
            & 53.80 & 74.31 & 86.94
            & 10.18 & 11.03 & 12.72 & 13.67 \\

            SEARLE-XL~\cite{searle}
            {\footnotesize\textcolor{gray}{(ICCV'23)}}
            & 24.24 & 52.48 & 66.29 & 88.84
            & 53.76 & 75.01 & 88.19
            & 11.68 & 12.73 & 14.33 & 15.12 \\

            ContextI2W~\cite{contexti2w}
            {\footnotesize\textcolor{gray}{(AAAI'24)}}
            & 25.60 & 55.10 & 68.50 & 89.80
            & -- & -- & --
            & -- & -- & -- & -- \\

            FTI4CIR~\cite{fti4cir}
            {\footnotesize\textcolor{gray}{(SIGIR'24)}}
            & 25.90 & 55.61 & 67.66 & 89.66
            & 55.21 & 75.88 & 87.98
            & 15.05 & 16.32 & 18.06 & 19.05 \\

            \cdashline{1-12}

            MagicLens~\cite{magiclens}
            {\footnotesize\textcolor{gray}{(ICML'24)}}
            & 30.10 & 61.70 & 74.40 & 92.60
            & 68.10 & 84.80 & 93.20
            & 29.60 & 30.80 & 33.40 & 34.40 \\

            \cdashline{1-12}

            CIReVL (GPT-3.5-Turbo)~\cite{cirevl}
            {\footnotesize\textcolor{gray}{(ICLR'24)}}
            & 24.55 & 52.31 & 64.92 & 86.34
            & 59.54 & 79.88 & 89.69
            & 18.57 & 19.01 & 20.89 & 21.80 \\

            LDRE (GPT-3.5-Turbo)~\cite{ldre}
            {\footnotesize\textcolor{gray}{(SIGIR'24)}}
            & 26.53 & 55.57 & 67.54 & 88.50
            & 66.43 & 80.31 & 90.05
            & 23.35 & 24.03 & 26.44 & 27.50 \\

            \hline

            MCL (OPT-2.7B)~\cite{mcl}
            {\footnotesize\textcolor{gray}{(ICML'24)}}
            & 23.28 & 54.17 & 67.16 & 90.05
            & 58.24 & 79.37 & 90.51
            & 14.55 & 15.79 & 17.38 & 18.27 \\

            MCL (OPT-6.7B)~\cite{mcl}
            {\footnotesize\textcolor{gray}{(ICML'24)}}
            & 24.15 & 55.98 & 69.21 & 90.82
            & 59.52 & 80.34 & 91.13
            & 14.14 & 16.13 & 17.88 & 18.82 \\

            MCL (LLaMA2-7B)~\cite{mcl}
            {\footnotesize\textcolor{gray}{(ICML'24)}}
            & 26.22 & 56.84 & 70.00 & 91.35
            & 61.45 & 81.61 & 91.93
            & 17.67 & 18.86 & 20.80 & 21.68 \\

            E5-V (LLaVA-NeXT-8B)~\cite{e5v}
            {\footnotesize\textcolor{gray}{(arXiv'24)}}
            & 33.90 & 64.12 & 75.88 & 93.54
            & 67.48 & 81.20 & 92.48
            & 18.48 & 19.21 & 20.95 & 21.83 \\

            \textbf{FiRE (BLIP-3-4B) (Ours)}
            & \textbf{43.33} & \textbf{74.02}
            & \textbf{83.51} & \textbf{95.83}
            & \textbf{73.01} & \textbf{88.38}
            & \textbf{94.94}
            & \textbf{31.03} & \textbf{32.08}
            & \textbf{34.40} & \textbf{35.50} \\

            \hline
            \hline

            \rowcolor{gray!15}
            {\footnotesize Ours vs. Dedicated ZS-CIR Model}
            & $\uparrow 13.23$
            & $\uparrow 12.32$
            & $\uparrow 9.11$
            & $\uparrow 3.23$
            & $\uparrow 4.91$
            & $\uparrow 3.58$
            & $\uparrow 1.74$
            & $\uparrow 1.43$
            & $\uparrow 1.28$
            & $\uparrow 1.00$
            & $\uparrow 1.10$ \\

            \rowcolor{gray!15}
            {\footnotesize Ours vs. Universal Retrieval Model}
            & $\uparrow 9.43$
            & $\uparrow 9.90$
            & $\uparrow 7.63$
            & $\uparrow 2.29$
            & $\uparrow 5.53$
            & $\uparrow 7.18$
            & $\uparrow 2.46$
            & $\uparrow 12.55$
            & $\uparrow 12.87$
            & $\uparrow 13.45$
            & $\uparrow 13.77$ \\

            \hline
        \end{tabular}%
    }
\end{table*}
\vspace{-1em}
\subsubsection{Implement Details}
MLLMs inherently consume significant memory~\cite{e5v}, and the long input embeddings involved in complex image retrieval tasks exacerbate this issue. Therefore, to mitigate memory consumption as much as possible, we used BLIP-3~\cite{blip3}, a more lightweight MLLM with only $4B$ parameters, as the model backbone for both fine-grained dataset generation and universal image retrieval. 

Regarding hyperparameters, for fine-grained dataset generation, we set the number of EOS tokens $M$ = 5 in Eq.(\ref{eq:feature}), the temperature coefficient $\tau$ of the image-text contrastive
loss in Eq.(\ref{eq:match_Loss}) as 0.01. The lower and upper thresholds $\theta_l$ and $\theta_h$ in Eq.(\ref{related_pairs}) are set to 0.6 and 0.83, respectively. For two-stage fine-tuning, we set $\tau'$ in Eq.(\ref{eq:infonce_loss})  as  0.01, temperature coefficients of the recall surrogate loss in Eq.(\ref{eq: recall loss}) as \( \tau_1 = 1 \) and \( \tau_2 = 0.01 \). Additionally, in Eq.(\ref{eq:total_loss}), we adopted both Recall@1 and Recall@5 to enhance the model's discriminative retrieval capability. 
The parameters are set as follows: $\mathcal{R}_s = [1, 5]$, with  $\beta_1 = 0.4$ and $\beta_2 = 0.15$. 
For all stages, we used the AdamW~\cite{AdamW} optimize for optimization.  

For both data generation and model fine-tuning, we froze the visual encoder and projection layers of the MLLM and only fine-tuned the LLM with LoRA~\cite{lora}, a lightweight parameter-efficient fine-tuning approach. In particular, we set the rank of the LoRA approximation to 64,  the \texttt{lora\_alpha} parameter to 128, and \texttt{lora\_dropout} parameter to 0.1. For deriving the MLLM encoder for relevant image pair identification in our data generation pipeline, we set the batch size as 16, the learning rate as $1\mathrm{e}{-4}$, and trained the model for 2 epochs. For the first-stage fine-tuning, we set the learning rate to $1\mathrm{e}{-4}$ and trained the model for 1 epoch, while for the second-stage fine-tuning, we set the batch size as 16, and the learning rate as $1\mathrm{e}{-4}$, and trained for 2 epochs.  
In addition, we leveraged DeepSpeed ZeRO-2~\cite{zero} for distributed training. We conducted all training using only 4 NVIDIA A100-40G GPUs. Notably, once the model is well trained, we keep the same checkpoint for evaluating our model on various tasks in a zero-shot setting.
\vspace{-1em}

\begin{table*}[!t]
    \centering
    \caption{
        \textbf{Performance comparison on FashionIQ in terms of
        $R@k$ (\%).}
        The best results are shown in bold.
        We also report the absolute performance improvements of our model
        over dedicated ZS-CIR models and universal retrieval models.
    }
    \label{tab:baseline_fashioniq}

    \resizebox{0.89\linewidth}{!}{%
        \begin{tabular}{l|cc|cc|cc|cc}
            \hline
            \multirow{2}{*}{Method}
            & \multicolumn{2}{c|}{Dresses}
            & \multicolumn{2}{c|}{Shirts}
            & \multicolumn{2}{c|}{Tops \& Tees}
            & \multicolumn{2}{c}{Avg.} \\
            \cline{2-3}
            \cline{4-5}
            \cline{6-7}
            \cline{8-9}

            & $\mathbf{R@10}$ & $\mathbf{R@50}$
            & $\mathbf{R@10}$ & $\mathbf{R@50}$
            & $\mathbf{R@10}$ & $\mathbf{R@50}$
            & $\mathbf{R@10}$ & $\mathbf{R@50}$ \\
            \hline
            \hline

            Pic2Word~\cite{pic2word}
            {\footnotesize\textcolor{gray}{(CVPR'23)}}
            & 20.00 & 40.20
            & 26.20 & 43.60
            & 27.90 & 47.40
            & 24.70 & 43.70 \\

            LinCIR~\cite{lincir}
            {\footnotesize\textcolor{gray}{(CVPR'24)}}
            & 20.92 & 42.44
            & 29.10 & 46.81
            & 28.81 & 50.18
            & 26.28 & 46.48 \\

            SEARLE-XL-OTI~\cite{searle}
            {\footnotesize\textcolor{gray}{(ICCV'23)}}
            & 21.57 & 44.47
            & 30.37 & 47.49
            & 30.90 & 51.76
            & 27.61 & 47.90 \\

            SEARLE-XL~\cite{searle}
            {\footnotesize\textcolor{gray}{(ICCV'23)}}
            & 20.48 & 43.13
            & 26.89 & 45.58
            & 29.32 & 49.97
            & 25.56 & 46.23 \\

            Context-I2W~\cite{contexti2w}
            {\footnotesize\textcolor{gray}{(AAAI'24)}}
            & 23.10 & 45.30
            & 29.70 & 48.60
            & 30.60 & 52.90
            & 27.80 & 48.93 \\

            FTI4CIR~\cite{fti4cir}
            {\footnotesize\textcolor{gray}{(SIGIR'24)}}
            & 24.39 & 47.84
            & 31.35 & 50.59
            & 32.43 & 54.21
            & 29.39 & 50.88 \\

            \cdashline{1-9}

            MagicLens~\cite{magiclens}
            {\footnotesize\textcolor{gray}{(ICML'24)}}
            & 25.50 & 46.10
            & 32.70 & 53.80
            & 34.00 & 57.70
            & 30.73 & 52.53 \\

            \cdashline{1-9}

            CIReVL (GPT-3.5-Turbo)~\cite{cirevl}
            {\footnotesize\textcolor{gray}{(ICLR'24)}}
            & 24.79 & 44.76
            & 29.49 & 47.40
            & 31.36 & 53.65
            & 28.55 & 48.57 \\

            LDRE (GPT-3.5-Turbo)~\cite{ldre}
            {\footnotesize\textcolor{gray}{(SIGIR'24)}}
            & 22.93 & 46.76
            & 31.04 & 51.22
            & 31.57 & 53.64
            & 28.51 & 50.54 \\

            \hline

            E5-V (LLaVA-NeXT-8B)~\cite{e5v}
            {\footnotesize\textcolor{gray}{(arXiv'24)}}
            & 23.75 & 47.45
            & 36.36 & 56.43
            & 35.29 & 57.47
            & 31.80 & 53.78 \\

            \textbf{FiRE (BLIP-3-4B) (Ours)}
            & \textbf{29.60} & \textbf{50.87}
            & \textbf{39.84} & \textbf{60.06}
            & \textbf{35.64} & \textbf{57.83}
            & \textbf{35.02} & \textbf{56.25} \\

            \hline
            \hline

            \rowcolor{gray!15}
            {\footnotesize Ours vs. Dedicated ZS-CIR Model}
            & $\uparrow 4.10$
            & $\uparrow 3.03$
            & $\uparrow 7.14$
            & $\uparrow 6.26$
            & $\uparrow 1.64$
            & $\uparrow 0.13$
            & $\uparrow 4.29$
            & $\uparrow 3.72$ \\

            \rowcolor{gray!15}
            {\footnotesize Ours vs. Universal Retrieval Model}
            & $\uparrow 5.85$
            & $\uparrow 3.42$
            & $\uparrow 3.48$
            & $\uparrow 3.63$
            & $\uparrow 0.35$
            & $\uparrow 0.36$
            & $\uparrow 3.22$
            & $\uparrow 2.47$ \\

            \hline
        \end{tabular}%
    }
\end{table*}

\subsubsection{Evaluation}
Following previous work~\cite{fti4cir,e5v,mcl}, we adopted the standard evaluation protocols to validate our approach on each dataset. 
For CIRR, we computed Recall at Rank $k$ (R@$k$) ($k = 1, 5, 10, 50$), as well as R$_{Subset}$@$k$ ($k = 1, 2, 3$) on the test split.
For FashionIQ, we adopted R@$k$($k = 10, 50$) for each category and reported the average metrics. For CIRCO, we adopted Mean Average Precision (mAP) as the metric, specifically mAP@$k$ ($k = 5, 10, 25, 50$). For Urban1K, Visual Dialog, COCO, and Flickr, we used R@$k$ ($k = 1, 5, 10$) as evaluation metrics. 

\subsection{On CIR Comparison }
For a comprehensive evaluation, we compared our method with not only the MLLM-based universal image retrieval models, including E5-V~\cite{e5v} and MCL~\cite{mcl}, but also several dedicated zero-shot CIR methods of three categories. 1) \textbf{Textual-inversion-based methods}, including Pic2Word~\cite{pic2word}, LinCIR~\cite{lincir}, SEARLE~\cite{searle}, Context I2W~\cite{contexti2w} and FTI4CIR~\cite{fti4cir}. These methods aim to pretrain a model to map images into pseudo word tokens, unifying the multimodal query into a token sequence, which can be processed by a pretrained VLP encoder for target image retrieval. 2) \textbf{Triplets-generation-based method}, i.e., MagicLens~\cite{magiclens}, which introduces a data generation pipeline to produce a large number of triplet samples for training a dual VLP encoder-based CIR model. 
Notably, the universal image retrieval model MCL and our proposed model also belong to this category. 
3) \textbf{LLM-based training-free methods}, including CIReVL~\cite{cirevl} and LDRE~\cite{ldre}, directly utilize strong LLMs to generate the target image caption based on the input reference image and modification text, thereby converting CIR to target text-to-image retrieval, which can be solved by VLP encoders. Notably, following E5-V~\cite{e5v}, we included the subset clothing category in the prompt, as \textit{``describe the <clothing category>''}, for promoting the model's performance on the FashionIQ dataset.

\begin{table}[t]
  \centering
  \caption{\textbf{Performance comparison on Visual Dialog and Urban1K with R@$k$(\%).} The best results are in boldface.}
  \vspace{-0.5em}
  \label{tab: baseline_dialog}
  \resizebox{1.0\linewidth}{!}{
  \normalsize
    \begin{tabular}{l|ccc|ccc}
    \hline
      \multirow{2}{*}{Method} & \multicolumn{3}{c|}{{Visual Dialog}}  & \multicolumn{3}{c}{{Urban1K}}\\
    \cline{2-7}
     & \textbf{R@$1$} & \textbf{R@$5$}& \textbf{R@$10$} & \textbf{R@$1$} & \textbf{R@$5$}& \textbf{R@$10$} \\
    \hline \hline
    CLIP~\cite{clip}          & 17.7 & 38.9 & 50.2&55.8&79.6&86.5 \\
    MCL(OPT-2.7B)~\cite{mcl}        & 25.6 & 51.9 & 65.2&–&–&– \\
    MCL(OPT-6.7B)~\cite{mcl}        & 27.2 & 51.0 & 64.0&–&–&– \\
    MCL(LLaMA2-7B)~\cite{mcl}      & 29.8 & 57.1 & 69.4&–&–&– \\  
    E5-V(LLaVA-NeXT-8B)~\cite{e5v} & 48.1 & 74.8 & 83.7&80.6 &93.9 &96.6 \\
    Long-CLIP~\cite{long-clip} & 35.4& 62.0& 72.7&86.1&96.4&98.1 \\
    \textbf{FiRE(BLIP-3-4B) (Ours)} & $\mathbf{54.9}$&$\mathbf{79.9}$ & $\mathbf{88.0}$ & $\mathbf{91.4}$&$\mathbf{98.0}$ & $\mathbf{99.2}$ \\
    \hline
    \end{tabular}
    }
      \vspace{-1.5em}
\end{table}
Table~\ref{tab:baseline_cirr} and Table~\ref{tab:baseline_fashioniq} present our results on CIRR, CIRCO, and FashionIQ. 
We directly used the results reported in the original papers of baselines, while we particularly reproduced the best universal image retrieval baseline E5-V with its public parameters,  using it as the main benchmark. 
Notably, the reported results of all baselines are based on the ViT-L/14 visual encoder, consistent with the encoder used in our MLLM. 
We also reported the improvements in our results compared to the existing best specialized ZS-CIR model and the leading MLLM-based universal image retrieval model, i.e., E5-V.
From these two tables, we have the following observations. 

1) Compared to LLM-based universal image retrievers, i.e., MCL and E5-V, our method, even with a more lightweight MLLM backbone, consistently shows performance improvement across all metrics. Specifically, the average of R@1 on CIRR, mAP@5 on CIRCO, and R@10 on FashionIQ, shows that our method improves by 8.4\% compared to the best universal baseline E5-V.
This shows the superior generalization capability of our model in various CIR contexts.

2) Compared to those dedicated CIR models, our universal model also consistently exhibits promising improvements across all metrics, demonstrating its superior multimodal context understanding capability.
Notably, although the LLM-based training-free methods (i.e., CIReVL and LDRE) use larger LLMs for inference, their retrieval performance remains lower than ours. This indicates the necessity of conducting proper LLM fine-tuning to enhance its multimodal context understanding capability. 
Additionally, compared to other triplets-generation-based methods (i.e., MCL and MagicLens), our method uses far fewer generated triplets (only $87K$) for model fine-tuning, whereas MCL and MagicLens utilize $2.7M$ and $36.7M$ generated triplets, respectively. This suggests the higher quality of our generated dataset, featuring more fine-grained differences and modification text that are closer to human annotations.

3) On FashionIQ, which involves more fine-grained modifications to garment details, our method shows more exceptional performance on the Dresses and Shirts subsets, compared to the  Tops\&Tees subset. One possible explanation is that compared to Tops\&Tees which involves various categories of tops, garments in Dresses and Shirts are more concentrated in a single clothing category, which requires the model to have stronger fine-grained reasoning capabilities. In these cases, the advantage of our model is highlighted.

\begin{table}[t]
  \centering
  \caption{\textbf{Performance comparison on COCO and Flickr with respect to R@$k$(\%).} The best results are in boldface.}
  \vspace{-0.5em}
  \label{tab: baseline_coco}
  \resizebox{8.5cm}{!}{
  \normalsize
    \begin{tabular}{l|ccc|ccc}
    \hline
      \multirow{2}{*}{Method} & \multicolumn{3}{c|}{{COCO}}  & \multicolumn{3}{c}{{Flickr}}\\
    \cline{2-7}
     & \textbf{R@$1$} & \textbf{R@$5$}& \textbf{R@$10$} & \textbf{R@$1$} & \textbf{R@$5$}& \textbf{R@$10$} \\
    \hline \hline
   CLIP~\cite{clip} & 35.4 & 60.1 & 70.2 & 68.7 & 90.6 & 95.2  \\ 
   MagicLens~\cite{magiclens} &44.3 &69.4 &78.3 & 72.5 &91.5 &95.2 \\
        Long-CLIP~\cite{long-clip} & 46.3 & 70.8 & 79.8 & 76.1 & 93.5 & 95.2 \\ 
        E5-V~\cite{e5v} & 52.0 & 76.5 & \textbf{84.7} & \textbf{79.5} & \textbf{95.0} & \textbf{97.6} \\ 
        \textbf{FiRE (Ours)} & \textbf{52.3} & \textbf{76.7} & $82.6$ & $76.2$ & $93.0$ & $95.5$ \\ 
    \hline
    \end{tabular}
    }
      \vspace{-1.0em}
\end{table}

\subsection{On Cross-modal Retrieval Comparison}
Apart from CIR, we also evaluated our model with three cross-modal image retrieval tasks, including two relatively complex tasks (i.e., long-text-to-image retrieval, and dialog-based image retrieval) and one standard task of short-text-to-image retrieval.

For the two complex retrieval tasks, to the best of our knowledge, there are no strong zero-shot baselines. Therefore, apart from the two universal image retrieval models (i.e., MCL and E5-V), we introduced Long-CLIP~\cite{long-clip}, a specialized zero-shot long-text retrieval model, as a baseline.
Table~\ref{tab: baseline_dialog} shows the performance comparison on the dialog-based image retrieval dataset Visual Dialog and long-text-to-image retrieval dataset Urban1K.
As can be seen, despite not being specifically trained for these two tasks, our method still shows exceptional performance. On the one hand, this verifies that conducting fine-grained context modeling inherently enhances the MLLM's understanding of complex queries. 
On the other hand, this demonstrates that fine-tuning MLLMs with the CIR task contributes significantly to improving complex image retrieval tasks. We attribute this to the fact that the CIR task itself is inherently complex with multimodal composite queries, which potentially enhances the model's complex query understanding capability. 

\begin{table}[t]
    \centering \caption{Ablation study on FashionIQ, CIRR, Visual Dialog, and Urban1K towards for key components of our method. }\vspace{-0.6em}
    \label{tab:ablation}
    \resizebox{1.0\linewidth}{!}{
    \begin{tabular}{l|cc|cc|cc|cc}
    \hline 
  \multirow{2}{*}{Method}  & \multicolumn{2}{c|}{\multirow{1}{*}{FashionIQ-Avg}} &\multicolumn{2}{c|}{\multirow{1}{*}{CIRR}}&\multicolumn{2}{c|}{\multirow{1}{*}{Visual Dialog}} & \multicolumn{2}{c}{\multirow{1}{*}{Urban1K}}  \\

    \cline{2-9}  &\textbf{R@$10$} & \textbf{R@$50$} & \textbf{R@$1$}  & \textbf{R@$5$} & \textbf{R@$1$}  & \textbf{R@$5$}  & \textbf{R@$1$}  & \textbf{R@$5$} \\
    \hline \hline

  w/-Img-LongCap    & 11.24 & 23.56 & 11.33 & 33.61 & 53.10 & 77.82 & \textbf{91.70} & \textbf{98.40} \\ 
  w/-OneStage & 32.72 & 53.60 & 41.67 & 72.35 & 53.21 & 78.99 & 87.90 & 96.20  \\

  w/o-FirstStage & 32.54 & 52.06 & 41.79 & 71.59 & 54.12 & 79.31 & 88.20 & 97.40 \\
  w/-ShortCap   & 32.21 & 52.63 & 42.33 & 72.72 & 53.45 & 78.97 & 88.90 & 97.20  \\
  w/o-RecallLoss   & 33.14 & 53.69 & 41.21 & 72.25 & 53.88 & 79.12 & 87.90 & 97.20 \\ 
   \hline
    \rowcolor{gray!15} \multicolumn{1}{c|}{\textbf{FiRE (Ours)}}    & \textbf{35.02} & \textbf{56.25} & \textbf{43.33}  & \textbf{74.02}  & \textbf{54.88} & \textbf{79.89} & 91.40 & 98.00 \\
    
    \hline
    \end{tabular} }
\end{table}

\begin{table}[t]
\vspace{-0.5em}
  \centering
  \caption{\textbf{Performance comparison of dataset on CIRR.} The best zero-shot results are in boldface.}
  \vspace{-0.5em}
  \label{tab: baseline_dataset}
   \resizebox{8.5cm}{!}{
  \normalsize
    \begin{tabular}{c|c|l|ccc|c}
  \hline
    \multirow{2}{*}{Supervision}  & \multirow{2}{*}{Dataset} & \multirow{2}{*}{Scale} & \multicolumn{3}{c|} {\textbf{R@$k$}} &\multirow{2}{*}{\textbf{R$_{Subset}$@$1$}}\\
    \cline{4-6}
    & & & {$k$ = $1$} & {$k$ = $5$}& {$k$ = $10$}&  \\
    \hline  \hline
   \multirow{3}{*}{Zero-Shot}   &MMC~\cite{mcl} & $2.7M$ &  21.74 & 51.54 & 65.33  & 49.28 \\
   & LaSCo~\cite{case} &$359.2K$ & 23.98 & 53.68 & 67.40  & 51.06 \\ 
   & FiGMaQ(Ours) & $87K$& $\mathbf{26.96}$ & $\mathbf{55.52}$ & $\mathbf{70.24}$  & $\mathbf{55.66}$ \\
    \hline \hline
   Supervised & CIRR & $28.2K$  & 28.17 & 57.51 & 71.74  & 58.77 \\
    \hline
    \end{tabular} }
   \vspace{-1em}
    \end{table}

Table~\ref{tab: baseline_coco} shows the performance comparison among different models on standard short-text-to-image retrieval datasets (i.e., COCO and Flickr), where we excluded MCL due to its missing results on these datasets, but instead incorporated CLIP and MagicLens that have reported the corresponding results. Since Long-CLIP has demonstrated outstanding performance in zero-shot cross-modal retrieval tasks, we also included it in this comparison. 
As can be seen, 
the LLM-based models, including E5-V and our model, outperform all the VLP-based models. This suggests the advantage of using the LLM over the VLP model as the encoder. Regarding the observation that our model slightly underperforms E5-V on Flickr, we attribute this to two key factors: 1) E5-V utilizes a larger LLM with 8B parameters as its backbone, while ours has only 4B parameters; and 2) E5-V is specifically trained with short text pairs, making it better suited for short-text-to-image retrieval scenarios, whereas our model is trained on long, fine-grained text pairs to enhance its complex context understanding capability. Nonetheless, our model still performs comparably to E5-V on the COCO dataset, demonstrating its effectiveness in simpler image retrieval tasks.
It is worth mentioning again that our model significantly outperforms E5-V on five datasets across various complex retrieval tasks.


\subsection{On Ablation Study}
To verify the importance of each component in our method, we
compared our method with its following derivatives.
\begin{itemize}
    \item \textbf{w/-Img-LongCap.} To explore the impact of using the CIR task for MLLM fine-tuning, we fine-tuned MLLM with the standard image-text alignment task instead of the two multimodal reasoning and retrieval tasks. Specifically, only \textless \emph{image, generated fine-grained caption}\textgreater ~pairs were used.  
     \item \textbf{w/-OneStage.} To explore the benefit of conducting two-stage fine-tuning, we mimicked MCL by simultaneously optimizing multimodal reasoning and retrieval tasks in one stage. 
      \item \textbf{w/o-FirstStage.} To explore the role of the fine-grained context reasoning-oriented fine-tuning phase, we disabled it. 
    \item \textbf{w/-ShortCap.} To validate the necessity of using fine-grained captions for multimodal context reasoning in the first fine-tuning stage, we replaced the fine-grained captions with coarse-grained captions~\cite{dqu} generated by BLIP-2~\cite{blip2}.  
    \item \textbf{w/o-RecallLoss.} To explore the effect of the recall surrogate losses, we fine-tuned the model without using them. 
\end{itemize}

\begin{figure}[t]
		\centering
		\includegraphics[width = \linewidth]{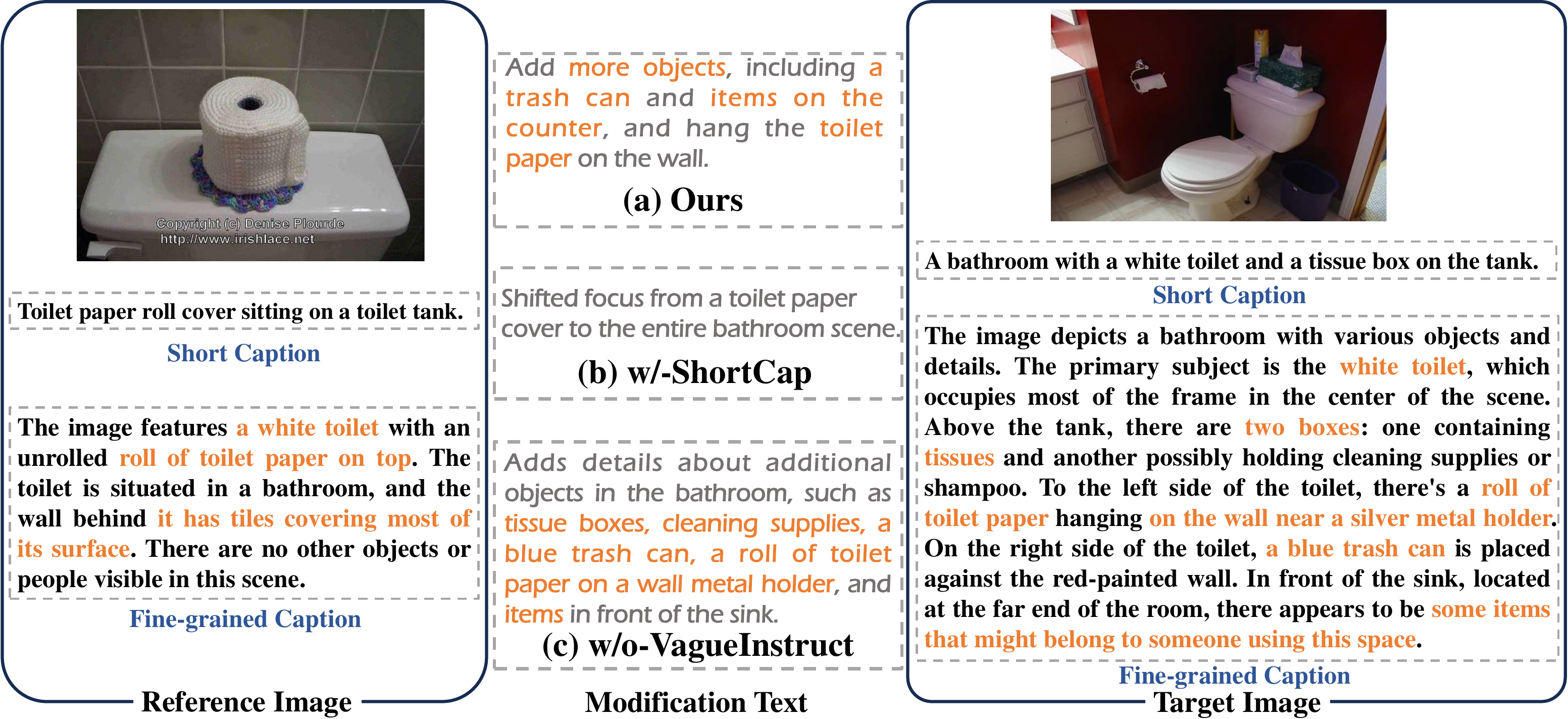}
        \vspace{-1em}
	    \caption{Modification text generated by our method and its two variants: w/-ShortCap and w/o-VagueInstruct. 
}\label{dataset}
\vspace{-1.0em}
        
\end{figure}

From Table~\ref{tab:ablation}, we have the following observations. 
1) w/-Img-LongCap shows significantly weaker performance than our method on CIR and dialog-based image retrieval tasks but has a slight advantage on the long-text-to-image retrieval task (Urban1K). This validates that fine-tuning with the CIR task and its derivative task (i.e., target caption generation task), due to their complexity and multimodal nature, better facilitates the enhancement of the MLLM's reasoning and contextual understanding abilities, compared to the simple cross-modal image-text alignment.  It is also reasonable that  w/-Img-LongCap specifically performs better on Urban1K, since its fine-tuning objective is totally aligned with the long-text-to-image retrieval task. 
2) Our method shows superior performance than both w/-OneStage and  w/o-FirstStage. This indicates the importance of sequentially conducting the fine-grained context reasoning-oriented fine-tuning and retrieval-oriented fine-tuning to ensure the model's universal performance on various image retrieval tasks. 
3) w/-ShortCap performs worse than ours on all tasks. This highlights the benefit of using fine-grained captions to improve the model's complex context reasoning capability. 
4) Compared to our method, w/o-RecallLoss performs poorly. This confirms the effect of recall surrogate losses in enhancing the model's discriminative query-target alignment capability. 
\vspace{-0.5em}
\subsection{On Dataset Comparison} 
\textbf{Quantitative Comparison.} To validate the quality of our dataset FiGMaQ, following~\cite{mcl}, we used it to train the classic CLIP encoder-based CIR model Combiner~\cite{clip4cir}. For comparison, we also adopted two large-scale publicly available CIR datasets: MMC (the dataset used by MCL) and LaSCo~\cite{case}, whose modification text is also automatically annotated by an LLM, to train the Combiner model. Table~\ref{tab: baseline_dataset} presents the zero-shot performance of Combiner models trained with different datasets, on the testing set of open-domain CIRR dataset. Notably, we also include the performance of Combiner in a supervised setting, where Combiner is trained with the training set of CIRR. As can be seen, despite having the smallest scale, our dataset achieves the best zero-shot performance. Meanwhile, we observed that the zero-shot performance of Combiner trained with our dataset is close to that trained in the conventional supervised setting. This verifies the high quality of our dataset, which stems from the following two reasons.
1) Our proposed fine-grained semantic filtering significantly reduces the inclusion of irrelevant pairs.
2) Our human-like fine-grained modification generation approach effectively mimics real human modifications, making the generated data more aligned with human annotation.

\textbf{Qualitative Comparison.} To gain a deep understanding of our generated dataset, we compared the modification text generated by our model and its two variants. 
1) \textbf{w/-ShortCap.} Generating modification text based on the coarse-grained image caption generated by BLIP-2~\cite{blip2} with LLaMA 3.1. 
2) \textbf{w/o-VagueInstruct.} Using LLaMA 3.1 to generate modification text based on the given pair of fine-grained image captions with the general instruction. Figure~\ref{dataset} illustrates the modification text comparison with an example, where we also provided the generated coarse-grained and fine-grained image captions for reference. 
As can be seen, compared to fine-grained captions, the  BLIP-2-generated coarse-grained image captions involve information loss on certain fine-grained details, leading to the LLM-generated modification text ``to the entire bathroom scene'' being too general and of little use for retrieving the target image. 
In addition, we find that the modification text generated by w/o-VagueInstruct is overly detailed, almost directly describing the target image. This can create biased triplets that hinder model fine-tuning and contrast with real-world scenarios, where users usually provide vague modifications focusing on key aspects of the image, rather than exactly detailing every element of the image. 
\begin{figure}[t]
		\centering
		\includegraphics[width = \linewidth]{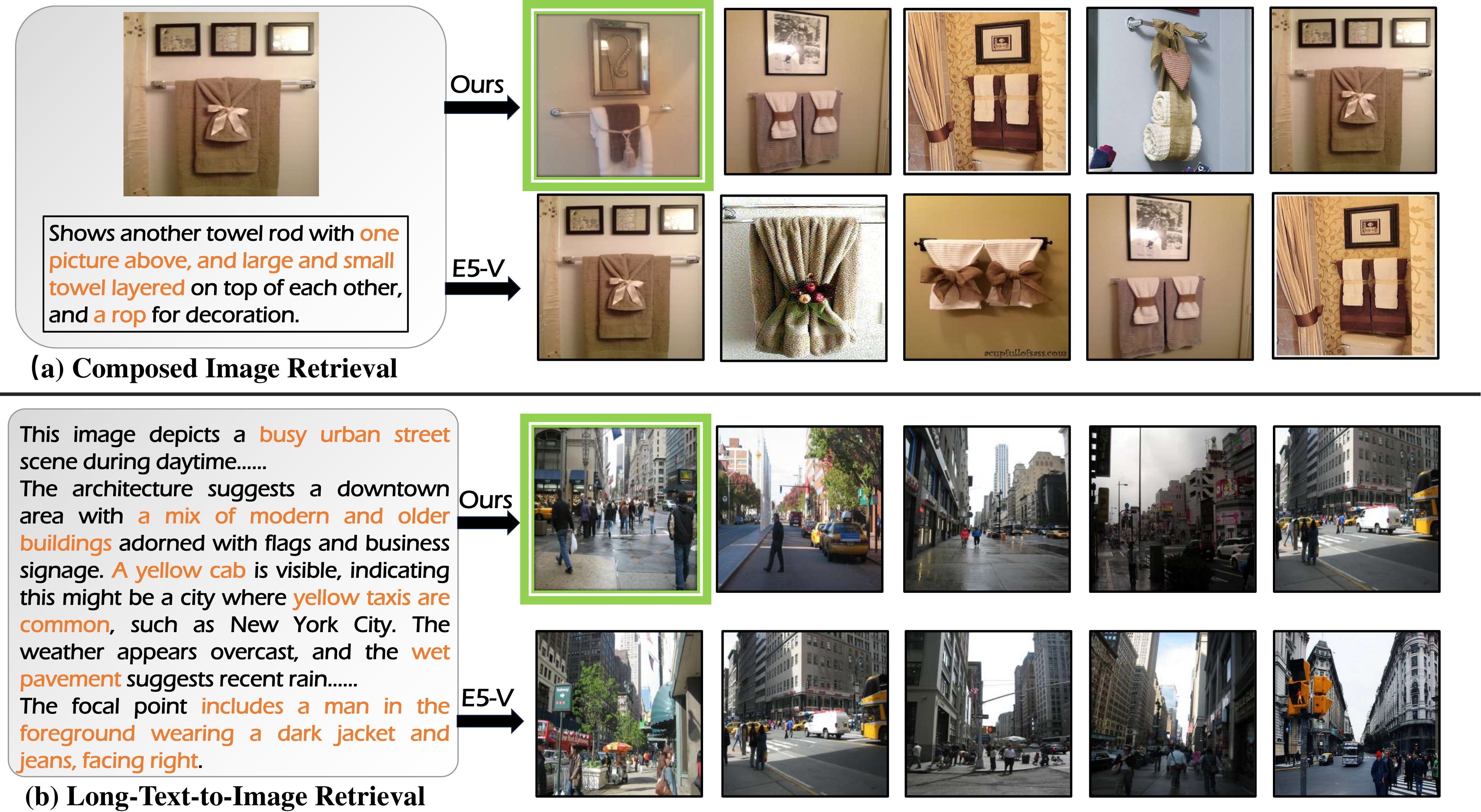}
        \vspace{-0.8em}
	    \caption{Illustration of CIR and Long-Text-to-Image Retrieval results, with ground-truth images highlighted in green boxes.}\label{casestudy}
        \vspace{-1.2em}
\end{figure}

\subsection{On Case Study}
\textbf{Retrieval Results}. Figure~\ref{casestudy} illustrates our retrieval results on tasks of CIR and long-text-to-image retrieval, compared with the best-performing universal image retrieval model, E5-V~\cite{e5v}. 
As illustrated in Figure~\ref{casestudy}(a), the given composed query requires fine-grained modifications to the reference image, involving multiple aspects, such as the subject quantity, spatial relationships, and subtle details (such as including a rop). For this case, our method accurately retrieves the ground truth in the first place, while E5-V fails, where its retrieved images cannot fully meet the modification demands, e.g., the quantity and spatial changes are not satisfied. As for the long-text-to-image retrieval case shown in Figure~\ref{casestudy}(b), E5-V performs significantly worse than ours. Specifically, E5-V primarily retrieves images that align with the general description of the query, which is insufficient to retrieve the correct image. In contrast, our method retrieves images that not only match the overall description but also satisfy fine-grained anchors in the query, such as a yellow taxi, rainy weather, or pedestrians with specified attire. These two cases demonstrate the effectiveness of our method in fine-grained context reasoning.

\section{Conclusion and Future Works}

In this work, we propose an automatic pipeline for constructing a fine-grained multimodal quintuple dataset and a novel two-stage fine-tuning strategy for MLLMs in complex image retrieval tasks. Using this pipeline, we create a large-scale dataset, FiGMaQ, to enhance fine-grained context modeling. Our strategy divides fine-tuning into two stages: (1) context reasoning-oriented fine-tuning and (2) retrieval-oriented fine-tuning, progressively improving context understanding and query-target alignment. Extensive experiments across five complex and two simple image retrieval tasks validate the effectiveness of our approach. Ablation and case studies further demonstrate the value of fine-grained fine-tuning. Future work will focus on expanding the dataset and exploring universal multimodal re-rankers for improved retrieval precision.

\section{Acknowledgments}
This work has been supported by the National Natural Science Foundation of China (No. 62376137, No. 624B2047, No.:62376140, No.:62276155 and No.:U23A20315), the Natural Science Foundation of Shandong Province (No. ZR2022YQ59). This work was also supported by Research Impact Fund (No.R1015-23),Collaborative Research Fund (No.C1043-24GF), Huawei (Huawei Innovation Research Program,Huawei Fellowship), Tencent (CCF-Tencent Open Fund, Tencent Rhino-Bird Focused Research Program), Alibaba (CCF-Alimama Tech Kangaroo Fund No. 2024002), Ant Group (CCF-Ant Research Fund), and Kuaishou.
\bibliographystyle{ACM-Reference-Format}
\newpage
\bibliography{mybib}
\end{document}